\UseRawInputEncoding
\documentclass{article}

\usepackage{microtype}
\usepackage{graphicx}
\usepackage{subcaption}
\usepackage{booktabs} % for professional tables
\usepackage{multirow} % for multirow cells in tables
\usepackage{hyphenat} % for \showhyphens command

\usepackage{hyperref}

\usepackage[preprint]{icml2026}

\usepackage{amsmath}
\usepackage{amssymb}
\usepackage{mathtools}
\usepackage{amsthm}

\usepackage[capitalize,noabbrev]{cleveref}

\theoremstyle{plain}

\theoremstyle{definition}

\theoremstyle{remark}

\usepackage[textsize=tiny]{todonotes}

\icmltitlerunning{Towards Explainable Adjudicative Variance: Quantifying Judicial Discretion via Gated Multi-Task Learning}

\begin{document}

\twocolumn[
  \icmltitle{Towards Explainable Adjudicative Variance: Quantifying Judicial Discretion via Gated Multi-Task Learning}



  \icmlsetsymbol{equal}{*}

  \begin{icmlauthorlist}
  \icmlauthor{Stanis{\l}aw S{\'o}jka}{tum}
  \icmlauthor{Felix Steffek}{cam}
  \icmlauthor{Matthias Grabmair}{tum}
\end{icmlauthorlist}

\icmlaffiliation{tum}{School of Computation, Information, and Technology, Technical University of Munich, Germany}
\icmlaffiliation{cam}{Faculty of Law, University of Cambridge}

\icmlcorrespondingauthor{Stanis{\l}aw S{\'o}jka}{stanley.sojka@tum.de}
\icmlcorrespondingauthor{Felix Steffek}{fs256@cam.ac.uk}
\icmlcorrespondingauthor{Matthias Grabmair}{matthias.grabmair@tum.de}

  \icmlkeywords{Machine Learning, ICML}

  \vskip 0.3in
]



\printAffiliationsAndNotice{}  % no special notice (required even if empty)

\begin{abstract}
    Legal outcome prediction must disentangle objective case facts from adjudicative context. Merit-based rulings rely on factual evidence while technical disposals may hinge on judicial discretion. We propose a Judge-Aware Gated Multi-Task Learning architecture that explicitly models this distinction. We introduce a fine-grained outcome taxonomy to supervise the encoder, enforcing a structural regularization that disentangles distinct semantic pathways. This granular legal curriculum enables our Gated Fusion mechanism to dynamically modulate reliance on judge identity. We evaluate our approach on 13,937 UK Employment Tribunal decisions. We benchmark our design against supervised fine-tuning (SFT) of a Gemma-4 26B-A4B backbone, in which judge identity and the taxonomy are injected as prompt tokens or autoregressive output targets. The two contextual signals compose only weakly when forced through a single autoregressive channel. In contrast, coupling a LoRA-adapted Gemma-4 encoder with our gated architecture defines a new state of the art on this benchmark while requiring an order of magnitude fewer trainable parameters than the generative SFT baselines, with gains concentrated on the most ambiguous and rarest outcome classes. Beyond accuracy, the architecture is interpretable; learned judge embeddings and calibration profiles localize the cases where adjudicative context drives the prediction. These results indicate that, for identity-conditioned classification of legal outcomes, the choice of conditioning interface dominates scale: differentiable structured composition yields more accurate, more parameter-efficient models than prompt-based composition over a substantially larger backbone.
\end{abstract}

\section{Introduction}

The rapid expansion of large-scale corpora and the advancements in Natural Language Processing (NLP) have positioned legal outcome prediction as a diagnostic instrument for dissecting the complex hierarchy of legal reasoning. These systems promise to democratize access to justice by estimating case prospects and streamlining judicial workflows. While formal rulings ensure that legal rights are upheld, providing transparency regarding likely outcomes is equally vital for facilitating amicable dispute resolution.

\begin{figure}[t!]
  \centering
  % trim format: left bottom right top
  \includegraphics[width=1.0\linewidth, trim=0mm 75mm 0mm 100mm, clip]{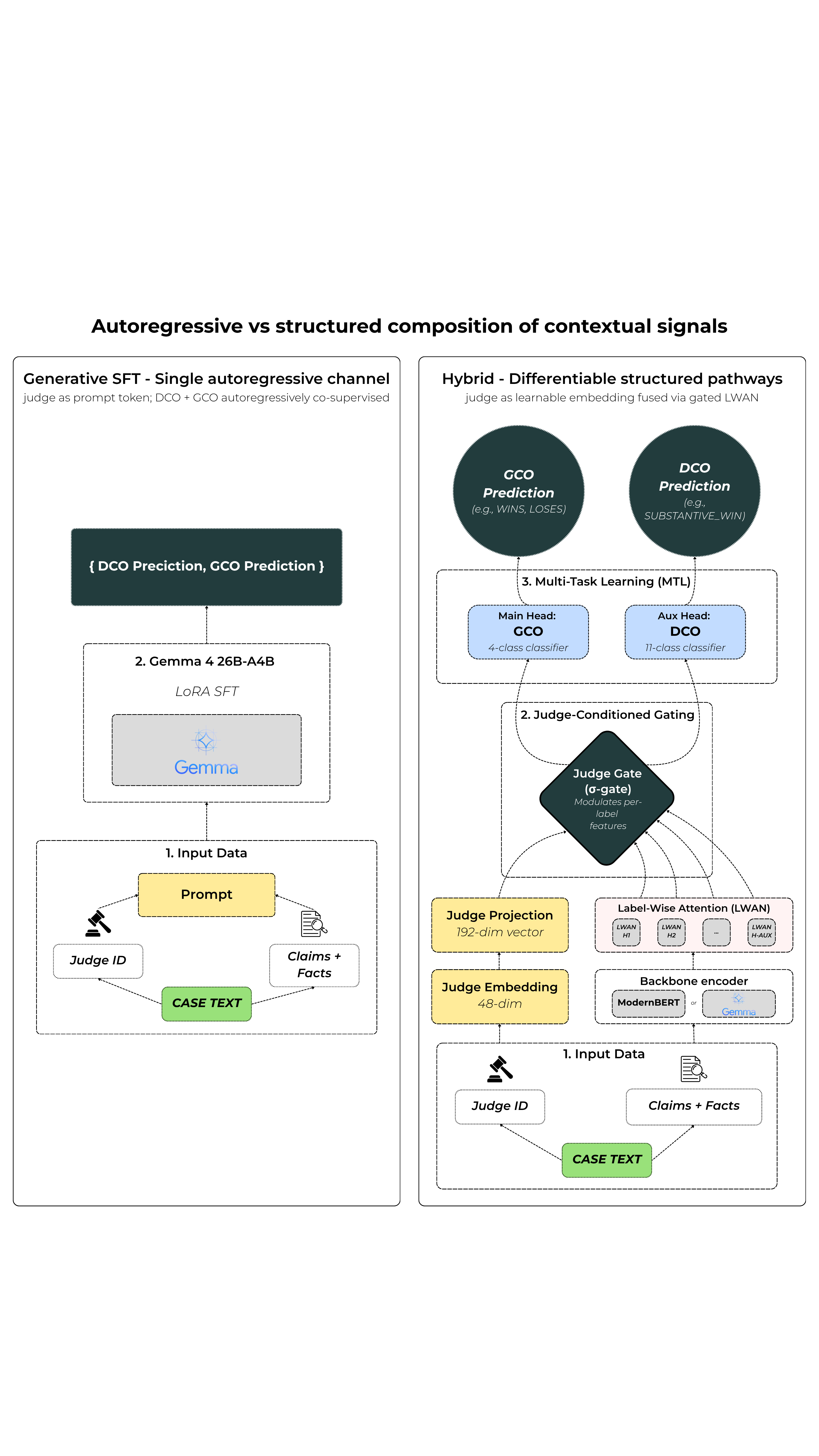}
  \caption{\textbf{Two routes for identity-conditioned hierarchical prediction.} \textit{Left:} generative supervised fine-tuning setup supplies judge identity as a prompt token and supervises the fine-grained DCO and coarse GCO labels through a single autoregressive output channel. \textit{Right:} the proposed hybrid architecture routes the same signals through differentiable components - a learned judge embedding fused into label-wise attention features by a gated mechanism, and parallel GCO/DCO classification heads. The two architectures share the same Gemma-4 26B-A4B backbone; only the conditioning interface differs.}
  \label{fig:architecture}
\end{figure}

Despite extensive research across various international and national jurisdictions, the AI-based prediction of UK court decisions - and specifically the nuanced domain of Employment Tribunals - remains relatively under-explored. While recent benchmarks have begun to address this gap \citep{xie2024clcuketdatasetbenchmarkingcase}, the unique interplay between procedural rules and fact-finding in the UK system presents challenges that standard text-classification models often fail to capture.

In this study, we focus on the cases of the \textbf{UK Employment Tribunal (UKET)} \cite{ostling2024cambridgelawcorpusdataset}, a high-volume jurisdiction where outcomes are not monolithic.
The UKET operates as a distinct adjudicative body, characterized by a mix of discretionary case management and strict procedural adherence. Claimants must satisfy procedural and substantive requirements within an adversarial framework, where the tribunal retains the power to dismiss claims at any stage for technical failures \citep{xie2024clcuketdatasetbenchmarkingcase}. Consequently, decisions frequently hinge on two distinct functional pathways:
\begin{enumerate}
    \item \textbf{Merit-Based Determinations:} Decisions where the judge evaluates the case facts and legal merits, even if there is no full oral hearing. This includes standard wins/losses and Default Judgments (which require a check of factual sufficiency).
    \item \textbf{Non-Merit-Based Disposals:} Decisions that terminate cases on technical grounds without engaging with the underlying facts. These hinge on jurisdiction, timelines, or non-compliance (e.g., Strike Outs or Out of Time rulings).
\end{enumerate}

Modeling this distinction is critical because the nature of judicial influence differs fundamentally between the two; a standard classifier that treats a strike out the same as a substantive win obscures the distinct mechanisms driving the decision.

While modern architectures, such as Multi-Task Learning (MTL), have proven advantageous for complex legal classifications, they remain predominantly judge-agnostic. By optimizing for a generalized judicial logic, these systems ignore the variance in individual decision-making patterns. Failing to explicitly incorporate the presiding judge as a dynamic variable limits the ability to account for the true drivers of case outcomes, especially in rulings heavily reliant on administrative discretion rather than objective case facts. 

Drawing inspiration from legal realism to address this gap, we posit that judicial identity is a decisive factor and employ two complementary mechanisms to model adjudicative variance: we incorporate judge identity embeddings to explicitly capture the discretion of the decision-maker, and we utilize fine-grained auxiliary supervision to improve the semantic representation of distinct legal pathways. 
Figure~\ref{fig:architecture} summarises our proposed processing pipeline.

The contributions of this paper are as follows:
\begin{itemize}
    \item We introduce a comprehensive legal analytics resource comprising a Detailed Case Outcome (DCO) taxonomy and judge identity. Our schema distinguishes between merit-based rulings (e.g., Substantive Win) and non-merit disposals (e.g., Strike Out) with unprecedented depth. This granular dataset extension opens new avenues for research into the interplay between case facts and judicial actors.

    \item We propose a novel Judge-Aware Gated Fusion mechanism. Unlike standard approaches that treat judge identity as a static feature, our architecture uses a dynamic gate to utilize judicial identity as a contextual anchor for ambiguous terminations. Crucially, by analyzing the model's sensitivity to counterfactual judge swaps and mapping the learned judicial embedding space, we provide a clear audit trail that disentangles objective case facts from judicial discretion. This ensures predictions are driven by structural legal understanding rather than opaque correlations.

    \item We provide the first controlled comparison of three regimes on the same UKET benchmark: a discriminative structured architecture built on ModernBERT, generative supervised fine-tuning of a Gemma-4 26B-A4B backbone that encodes judge identity and DCO supervision as prompt and output tokens, and a hybrid that keeps the discriminative structured head but routes its features through the same Gemma-4 encoder. We show that prompt-level composition of judge identity and fine-grained taxonomic supervision is \emph{sub-additive}, while the same two signals compose additively when routed through differentiable architectural components.

    \item We achieve a new performance frontier with intrinsic transparency. Evaluated on 13,937 UK Employment Tribunal decisions, the hybrid architecture (LoRA-adapted Gemma-4 encoder + LWAN + judge-gated fusion + MTL head) reaches 65.21 macro-F1, improving by $+5.1$ points over the strongest generative SFT variant, while training roughly an order of magnitude fewer parameters than the generative baselines. The gains concentrate on the fuzzy and rare outcome classes where structured composition should matter most, demonstrating that interpretability is downstream of the differentiable conditioning interface.
\end{itemize}

\section{Related Work}

\paragraph{Legal Judgment Prediction across Jurisdictions}
The task of Legal Judgment Prediction (LJP) has been studied in various settings \cite{cui2022surveylegaljudgmentprediction}, ranging from the US Supreme Court \citep{katz2017general} to international courts such as the European Court of Human Rights (ECHR) \citep{Aletras2016PredictingJD, chalkidis-etal-2019-neural, chalkidis-etal-2022-lexglue, t-y-s-s-etal-2023-zero}. Research has also extended to national jurisdictions including the Indian Supreme Court \cite{malik-etal-2021-ildc}, Chinese Criminal Courts \cite{luo-etal-2017-learning, Zhong2018LegalJP, Li2019MANNAM}, and the Federal Supreme Court of Switzerland \citep{niklaus-etal-2021-swiss, Trautmann2022LegalPE}, and recently, large-scale benchmarks for Korean legal language understanding \cite{hwang2022}.

\paragraph{Methodological Approaches to Context Awareness}
Earlier approaches for LJP were often limited to classifiers that conflated distinct legal pathways and were vulnerable to shallow surface signals \cite{ santosh-etal-2022-deconfounding, t-y-s-s-etal-2024-towards-supporting}. To improve context awareness, previous studies have leveraged strategies such as precedents \cite{wu-etal-2023-precedent, t-y-s-s-etal-2024-incorporating}, contrastive learning \cite{t-y-s-s-etal-2023-leveraging, gan-etal-2023-exploiting}, the entanglement between charges and articles \cite{dong2021}, or auxiliary information like court composition and chronological trends \cite{katz2017general, niklaus-etal-2021-swiss}. Crucially, \citet{t-y-s-s-etal-2024-towards} demonstrated that higher prediction accuracy does not guarantee better explainability or fairness, underscoring the need for models that explicitly disentangle decision factors.

\paragraph{Multi-Task Learning and Judicial Attributes}
Multi-task learning (MTL) has established itself as a prevalent framework in LJP \cite{Wang2025TheRA}. By formulating learning as a multi-objective optimization, MTL frameworks can enforce logical consistency between laws, charges, and penalties \cite{Zhong2018LegalJP}. Most recently, \citet{li-etal-2025-legal} advanced this by incorporating label-level knowledge definitions to enhance task synergy. However, these approaches remain predominantly judge-agnostic. While quantitative political science has long tracked judicial ideology through pre-appointment editorials \cite{Segal_1989} or voting records \cite{Martin_Quinn_2002}, and recent work using traditional machine learning has highlighted the importance of judge-specific variables \cite{medvedeva2020using, zambrano2025judgevariablechallengingjudgeagnostic}, few deep learning architectures explicitly model the presiding judge as a dynamic variable. Our work bridges this gap by combining the structural benefits of MTL with explicit judge-identity modeling to capture adjudicative variance.

\paragraph{Generative Fine-Tuning of Legal LLMs and LLM-as-Encoder}
A complementary line of work fine-tunes large generative LLMs directly for legal classification and judgment prediction. Lawma \citep{dominguezolmedo2024lawma} fine-tunes Llama-3 on 260 US Supreme Court and Appeals Court tasks via full supervised fine-tuning; domain-specific systems such as DISC-LawLLM \citep{yue2023disclawllm} and LegiLM \citep{deng2024legilm} follow a similar template of instruction tuning for legal classification, retrieval, and compliance, and recent two-stage pipelines such as Unilaw-R1 \citep{li2025unilawr1} extend it with reinforcement-style post-training. These systems treat legal context - including judge or court identity when available - as part of the input prompt, and supervise the output as a generated token sequence. A separate line of work re-purposes decoder-only LLMs as text encoders for downstream classification \citep{behnamghader2024llm2vec, springer2024echoembeddings, skean2025layerbylayer, li2025moeembedding}, exposing intermediate representations and mixture-of-experts router activations as features for task-specific heads rather than relying on autoregressive generation.

\section{Method: Modeling Adjudicative Variance via Merit-Based and Non-Merit-Based Pathways}
\label{sec:method}

To disentangle the latent behavioral signals driving the rulings, we propose an architecture designed to explicitly model the interplay between legal text, outcome granularity, and judicial context. Our approach begins with a discriminative baseline built on the ModernBERT encoder \citep{warner-etal-2025-smarter}. To distinguish technical disposals from merit-based decisions, we first define a DCO taxonomy, employing an LLM-assisted pipeline to extract 11 fine-grained categories from the training corpus. These granular definitions serve as the semantic prototypes for a Label-Wise Attention Network (LWAN), which maps the document embedding into distinct representation subspaces for each DCO label. To enforce logical consistency across these independent heads, we implement a Multi-Task Learning objective that constrains the model to align coarse decisions with their fine-grained constituents. Finally, we utilize these refined outcome features to drive a \textbf{Judge-Aware Gated Fusion} mechanism; this module dynamically modulates reliance on judicial embeddings, activating them only when the model detects discretionary or procedural nuance. 
\paragraph{Task definition.}
\label{def:outcome_labels}
We model legal outcome prediction as a four-class classification task. Following the taxonomy of \citet{xie2024clcuketdatasetbenchmarkingcase}, the coarse General Case Outcome (GCO) label, denoted as $y \in \mathcal{Y}_\mathrm{GCO}$, takes one of four values: \textit{Claimant Wins}, \textit{Claimant Loses}, \textit{Claimant Partly Wins}, and \textit{Other}. All models in this paper are evaluated on this label space; the fine-grained DCO labels introduced in \S\ref{sec:dco} serve as auxiliary supervision only.

\subsection{Discriminative Baseline}
\label{sec:baseline}

Our discriminative architecture uses a ModernBERT encoder \citep{warner-etal-2025-smarter} as the textual backbone; its 8{,}192-token native context allows full ingestion of the concatenated facts and claims without truncation \citep{sargeant-etal-2025-detecting}. We deliberately separate the encoder from the conditioning interface: judicial identity enters through a learned embedding fused by a gated mechanism (\S\ref{sec:judge-fusion}), and fine-grained outcome supervision enters through an auxiliary head (\S\ref{sec:mtl}), not through prompt tokens or generative output channels. This separation is the architectural commitment whose role we isolate by contrasting it against generative fine-tuning (\S\ref{sec:g-track}) and the hybrid B-track (\S\ref{sec:b-track}).

\subsection{DCO Taxonomy}
\label{sec:dco}

Building on prior research that utilizes the standard four-way classification for this dataset, we explore a complementary perspective. We introduce a deep semantic representation through a fine-granular taxonomy.

To capture this latent structure, we developed a taxonomy of 11 \textit{Detailed Case Outcomes} in consultation with two domain experts (UK labor law scholars). We then applied these predefined categories at scale across the training corpus using an LLM-assisted extraction pipeline \citep{openai2025gpt5}. Crucially, we treat these automated labels as noisy auxiliary targets rather than absolute ground truth, leveraging them as a structural regularizer that forces the encoder to learn robust semantic distinctions.

In this paper, we adopt a functional taxonomy that focuses on whether the judge considers the merits of the case, rather than relying on jurisdiction-specific doctrinal labels. The DCO schema distinguishes between \textbf{Merit-Based Outcomes}, \textbf{Non-Merit-Based Disposals} and \textbf{Mixed Outcomes} such as Claimant Partly Wins, which may involve a combination of merit-based findings and technical disposals (see Appendix~\ref{app:dco_taxonomy} for taxonomy details).
These fine-grained labels serve two purposes: they define the semantic prototypes for the attention mechanism and act as auxiliary supervision targets to regularize the shared encoder.

\subsection{Label-Wise Attention for Outcome-Specific Evidence}
\label{sec:lwan}

Different legal outcomes are often supported by different parts of the case description. To capture this structure, we adopt a Label-Wise Attention Network \cite{mullenbach-etal-2018-explainable}, demonstrated to be highly effective for large-scale legal classification~\cite{chalkidis-etal-2019-large, chalkidis-etal-2020-empirical}.

We observe that initializing the label-wise mechanism with a globally pooled document representation yields superior performance. We attribute this to the strength of the underlying ModernBERT encoder: unlike the RNN-based encoders used in early LWAN literature~\cite{chalkidis-etal-2019-large}, ModernBERT already leverages deep self-attention to contextualize dependencies across the entire sequence. Consequently, applying token-level attention on top of these rich embeddings introduces redundancy and increases sensitivity to local noise.

We compare the ModernBERT embedding to a set of $K$ learned outcome prototypes $\{l_k\}_{k=1}^K$, where $K$ corresponds to the number of fine-grained DCO labels used in the model. For each prototype, we compute a compatibility score, which is normalized across labels to obtain attention weights. The case embedding is projected into an attention feature space and reweighted to form per-label features. This taxonomy-specific representation allows the model to associate specific semantic signals with different outcome categories.

\subsection{Judge-Aware Gated Fusion}
\label{sec:judge-fusion}

A naive concatenation of judge embeddings treats judicial identity as a static feature, failing to capture its context-dependent role. We instead adopt a \textit{Gated Fusion} module that acts as a dynamic reliability switch: it ``consults'' the judge embedding when the textual evidence is discretionary - where discretion denotes adjudicative variance arising from differing judicial styles of case management - while suppressing it for clear-cut outcomes.

Gating mechanisms have been extensively validated for regulating information flow across heterogeneous modalities \cite{arevalo2017gatedmultimodalunitsinformation, dauphin2017languagemodelinggatedconvolutional, LIM20211748, shihata2025gatedrecursivefusionstateful}; we adapt the logic into a static conditional block that treats judicial identity as a latent control variable.

Let $\mathbf{u}_j \in \mathbb{R}^{d_j}$ denote a learned embedding for judge $j$. We project this embedding into the attention feature space to obtain $\mathbf{v}_j$. For each outcome label $k$, we compute a scalar gate activation $g_k \in (0, 1)$:
\begin{equation}
    g_k = \sigma(\mathbf{W}_g [\mathbf{c}_k; \mathbf{v}_j] + b_g)
\end{equation}
where $\mathbf{c}_k$ is the text-derived feature for outcome $k$. We then fuse the judge signal via a residual connection modulated by this gate:
\begin{equation}
    \hat{\mathbf{c}}_k = \mathbf{c}_k + \tanh(\mathbf{W}_f [\mathbf{c}_k; g_k \odot \mathbf{v}_j] + b_f)
\end{equation}
This design allows the model to learn outcome-specific reliance on judicial context---for example, heavily weighting judge identity for a \textit{Substantive Win} while ignoring it for an \textit{Out of Time} dismissal.

\subsection{Auxiliary Outcome Supervision and Multi-Task Learning}
\label{sec:mtl}

To expose merit-based and non-merit-based latent structure during training, we augment supervision with the fine-grained DCO labels introduced in \S\ref{sec:dco}.

Let $z^{(g)} \in \mathcal{Z}^{(g)}$ denote an auxiliary outcome label at granularity $g$, where $|\mathcal{Z}^{(g)}| \in \{6,8,11\}$ in our experiments. We attach an auxiliary classifier to the shared representation and predict the fine-grained outcome category
\begin{equation}
  p\bigl(z^{(g)} \mid x\bigr) = \mathrm{softmax}\bigl(W_{z^{(g)}}\, \phi(x) + b_{z^{(g)}}\bigr),
\end{equation}
where $\phi(x)$ denotes a pooled representation derived from the fused label-wise features. Training this auxiliary head induces a standard cross-entropy loss. The main outcome prediction and auxiliary task are trained jointly using a weighted multi-task objective
\begin{equation}
  \mathcal{L} = \mathcal{L}_{\mathrm{main}} + \alpha\, \mathcal{L}_{\mathrm{aux}}^{(g)},
\end{equation}
where $\alpha \ge 0$ controls the contribution of auxiliary supervision. We treat both the auxiliary granularity $g$ and the weight $\alpha$ as controlled variables and study their effect through cross-granularity and $\alpha$-sensitivity analyses (see Appendix~\ref{app:b1_encoder}).

\subsection{Generative Supervised Fine-Tuning Baselines}
\label{sec:g-track}

To probe whether a substantially larger autoregressive language model can absorb the same fine-grained outcome structure and judge identity signal through prompting and output formatting alone, we evaluate a family of \emph{generative} baselines (G-track) built on the Gemma-4 26B-A4B mixture-of-experts backbone \citep{gemmateam2026gemma4}. All G-track variants are fine-tuned with LoRA \citep{hu2022lora}, yielding nearly an order of magnitude more than the discriminative head described in \S\ref{sec:judge-fusion}--\S\ref{sec:mtl} (see Appendix~\ref{app:training} for full hyperparameters).

We organize the G-track around a $2{\times}2$ factor of two binary conditioning choices: whether judge identity is supplied as a prompt token, and whether the model is supervised to co-emit the DCO label alongside the coarse GCO outcome in a JSON envelope. Concretely, \textbf{G1} provides no judge token and GCO-only output; \textbf{G2} adds DCO+GCO co-prediction; \textbf{G3} adds judge identity as a prompt token with GCO-only output; \textbf{G4} combines both. We additionally report a zero-shot baseline \textbf{G0} and two diagnostic variants \textbf{G4-H1}/\textbf{G4-H2} to rule out scale and formatting confounds.

\subsection{Hybrid Architecture with a Gemma-4 Encoder}
\label{sec:b-track}

The G-track tests whether prompt-level conditioning is sufficient. We complement it with a \emph{hybrid} family (B-track) that re-uses the structured discriminative head from \S\ref{sec:lwan}--\S\ref{sec:mtl} but replaces the ModernBERT encoder with the same Gemma-4 26B-A4B backbone, so that the only axis of variation against the G-track is the conditioning interface.

\paragraph{B1: frozen encoder + structured head.}
B1 freezes the Gemma-4 backbone and trains only the LWAN head with judge-gated fusion and the MTL objective. Motivated by recent evidence that intermediate representations of decoder-only LLMs outperform their last-layer counterparts on classification probes \citep{skean2025layerbylayer}, we extract features from an interior transformer layer ($L_{15}$, $\approx{}50\%$ depth); the broader line of work on converting decoder-only LLMs into text encoders \citep{behnamghader2024llm2vec} motivates this hybrid design. Because the causal backbone restricts each token to a prefix-only context, we adopt the \emph{echo embedding} readout \citep{springer2024echoembeddings} to recover bidirectional context without retraining the backbone. The encoder pass is run once and its features cached; the structured head then trains from the cache. Layer selection details, the echo-embedding mechanism, and the MoE router feature concatenation are described in Appendix~\ref{app:b1_encoder}.

\paragraph{B2: end-to-end LoRA encoder + structured head.}
B2 unfreezes the Gemma-4 encoder via LoRA on the same projection set as the G-track and trains it jointly with the LWAN head, judge-gated fusion, and MTL objective. B2 shares the G-track's broad signal palette---case text, judge identity, and fine-grained DCO supervision---but routes judge identity through a learned embedding fused via the gated attention of \S\ref{sec:judge-fusion} and DCO supervision through a parallel classification head, isolating the \emph{conditioning interface} as the variable we read off in the results.

\section{Evaluation and Results}
\label{sec:results}

We evaluate our approach on the UKET outcome prediction task using a series of controlled ablation studies designed to isolate the contribution of each major architectural component. We report \textbf{Macro-F1} scores to assess performance on minority and legally rare outcomes, and \textbf{Weighted-F1} scores to reflect overall predictive quality under class imbalance.

\subsection{Dataset and Preprocessing}

The experimental dataset is derived from the \texttt{CLC-UKET\textsubscript{pred}} corpus \citep{xie2024clcuketdatasetbenchmarkingcase}, comprising UK Employment Tribunal decisions from 2011 to 2023. After resolving judicial identities for our modeling requirements, the final dataset contains 13,937 cases. Input documents were encoded using ModernBERT, processing a concatenated string format of ``Facts: [facts] \textbackslash n\textbackslash n Claims: [claims]'' to capture both the factual background and the nature of the dispute. Detailed dataset statistics, the complete filtering pipeline, and our judge-aware data splitting strategy are described in Appendix~\ref{app:dataset_statistics}.

\subsection{D-Track: Discriminative Architecture Ablation}
\label{sec:results-dtrack}

Table~\ref{tab:dtrack-ablation} reports results for progressively richer discriminative variants on the UKET test set. Starting from a frozen-encoder baseline, introducing MTL with a CLS-based representation yields a substantial improvement in both reported scores. Replacing CLS pooling with LWAN further improves Macro-F1, indicating that outcome-specific representations better capture heterogeneous legal reasoning patterns. Removing judge information from the LWAN+MTL model leads to a consistent degradation, demonstrating that judge context provides complementary signal beyond textual evidence alone.

\begin{table}[t]
\centering
\scalebox{0.85}{
\begin{tabular}{lcc}
\hline
\textbf{Model / Setting} & \textbf{Macro-F1} & \textbf{Weighted-F1} \\ \hline
\multicolumn{3}{l}{\textit{Architectural progression}} \\
Baseline (CLS) & 43.24 & 60.90 \\
CLS MTL Fusion & 49.93 & 62.56 \\
LWAN Single-Task Fusion & 50.59 & 62.64 \\
LWAN MTL (no judge) & 49.59 & 62.11 \\ \hline
\multicolumn{3}{l}{\textit{Judge fusion mechanism (LWAN+MTL)}} \\
Concat Fusion & 43.41 & 61.54 \\
Cross-Attention Fusion & 48.81 & 63.39 \\
\textbf{Gated Fusion (D4c)} & \textbf{52.66} & \textbf{64.46} \\ \hline
\multicolumn{3}{l}{\textit{Auxiliary task granularity (LWAN+MTL+Gated)}} \\
6-class DCO & 49.96 & 62.60 \\
8-class DCO & 50.66 & 64.63 \\
\textbf{11-class DCO (D4c)} & \textbf{52.66} & \textbf{64.46} \\ \hline
\end{tabular}}
\caption{D-track ablation on UKET outcome prediction with the ModernBERT-based discriminative architecture. The best D-track model (D4c) combines LWAN, 11-class DCO multi-task supervision, and judge-aware gated fusion.}
\label{tab:dtrack-ablation}
\end{table}

\paragraph{Efficacy of gated modulation.}
We benchmark our proposed gated fusion (\S\ref{sec:judge-fusion}) against \textit{concatenation fusion}, which applies judge embeddings uniformly, and \textit{Multi-Head Cross-Attention} \cite{Vaswani2017AttentionIA}, which enables outcomes to dynamically query judge characteristics. Gated fusion substantially outperforms both, demonstrating that selectively modulating influence at the outcome level is superior to uniform conditioning, and the learned gates provide a valuable interpretable signal for analyzing outcome-specific judge dependence.

Stratifying the test set by gate activation magnitude reveals a semantically meaningful pattern: accuracy increases monotonically with gate influence, straightforward technical terminations (\textit{strike\_out\_non\_pursuit}) concentrate in the lowest quartile, and \textit{default\_judgment} expands dramatically in the highest---consistent with the intuition that procedurally discretionary rulings require the model to ``check'' judge identity. Full quartile-level accuracy and DCO composition plots are given in Appendix~\ref{app:gate_influence}.

\paragraph{Legal nuance as regularization.}
Increasing the granularity of the auxiliary DCO labels leads to consistent improvements. Moving from a 6-class to an 8-class auxiliary task yields moderate gains, while the full 11-class DCO supervision provides the strongest performance, supporting our hypothesis that \emph{legal nuance acts as a form of regularization}: richer outcome structure during training improves generalization on the coarse four-way prediction task, consistent with observations in \citet{article, Sanh2018AHM, Elnaggar2018MultiTaskDL, Dou2019DomainGV}.

We additionally verified that the structured head does not collapse into redundant prototypes: pairwise Pearson correlations between the $11$ learned LWAN label-embedding vectors hover near zero (average $|\rho|<0.05$), with no off-diagonal entries above $0.9$. The attention mechanism therefore queries independent semantic subspaces, avoiding the head-collapse failure mode documented for multi-head attention \cite{voita-etal-2019-analyzing, kovaleva-etal-2019-revealing}.

\subsection{Cross-Track Comparison: G-Track and B-Track}
\label{sec:results-crosstrack}

Table~\ref{tab:crosstrack} compares the best D-track model (D4c) against the eight Gemma-4 26B-A4B variants from the G-track and B-track; only the conditioning interface and trainable head differ across the Gemma-4 runs.

\begin{table}[t]
\centering
\scalebox{0.78}{
\begin{tabular}{llcc}
\hline
\textbf{Track} & \textbf{Model} & \textbf{Macro-F1} & \textbf{Weighted-F1} \\ \hline
D & D4c (best D-track) & 52.66 & 64.46 \\ \hline
G & G0 (zero-shot) & 49.18 & 61.73 \\
G & G1 (vanilla SFT) & 57.22 & 72.30 \\
G & G2 (+DCO co-pred.) & 59.17 & 72.12 \\
G & G3 (+judge prompt) & 60.07 & 72.88 \\
G & G4 (+judge $+$ DCO) & 60.12 & 73.26 \\
G & G4-H1 ($r{=}64$) & 59.67 & 71.91 \\
G & G4-H2 (GCO-first) & 59.54 & 72.93 \\ \hline
B & B1 (frozen enc. $+$ head) & 58.05 & 69.48 \\
B & \textbf{B2 (LoRA enc. $+$ head)} & \textbf{65.21} & \textbf{75.55} \\ \hline
\end{tabular}}
\caption{Cross-track comparison on UKET. D-track: best ModernBERT-based discriminative model. G-track: Gemma-4 26B-A4B with LoRA SFT under different prompt and target configurations. B-track: Gemma-4 26B-A4B used as encoder for the same structured head (B1 frozen, B2 jointly fine-tuned via LoRA).}
\label{tab:crosstrack}
\end{table}

\paragraph{G-track: sub-additive composition.}
While judge conditioning and detailed case outcome (DCO) co-prediction individually improve generative fine-tuning, their prompt-level composition is strikingly sub-additive (Table~\ref{tab:crosstrack}). Combining both signals (G4) merely plateaus at the performance of judge-conditioning alone (G3), falling well short of additive expectations. Diagnostic variants confirm this plateau is not an artifact of LoRA capacity or JSON formatting. Instead, the bottleneck is structural: the single autoregressive channel forces near-perfect consistency between coarse and fine predictions, effectively collapsing the independent, parallel multi-task pathways that a structured architecture naturally preserves.

\paragraph{B-track: restoring compositional structure.}
B1, which freezes the Gemma-4 backbone and trains only the structured head, already beats vanilla generative SFT (G1) while training far fewer parameters; the rare-class signal it surfaces shows that useful adjudicative structure is already present in Gemma-4's intermediate representations. B2, which adds LoRA on the encoder and trains it jointly with the head, sets the new frontier at \textbf{65.21 macro-F1}, indicating that the binding constraint is the conditioning interface rather than backbone scale or LoRA capacity. 

\begin{figure}[t]
        \centering
        \includegraphics[width=\linewidth]{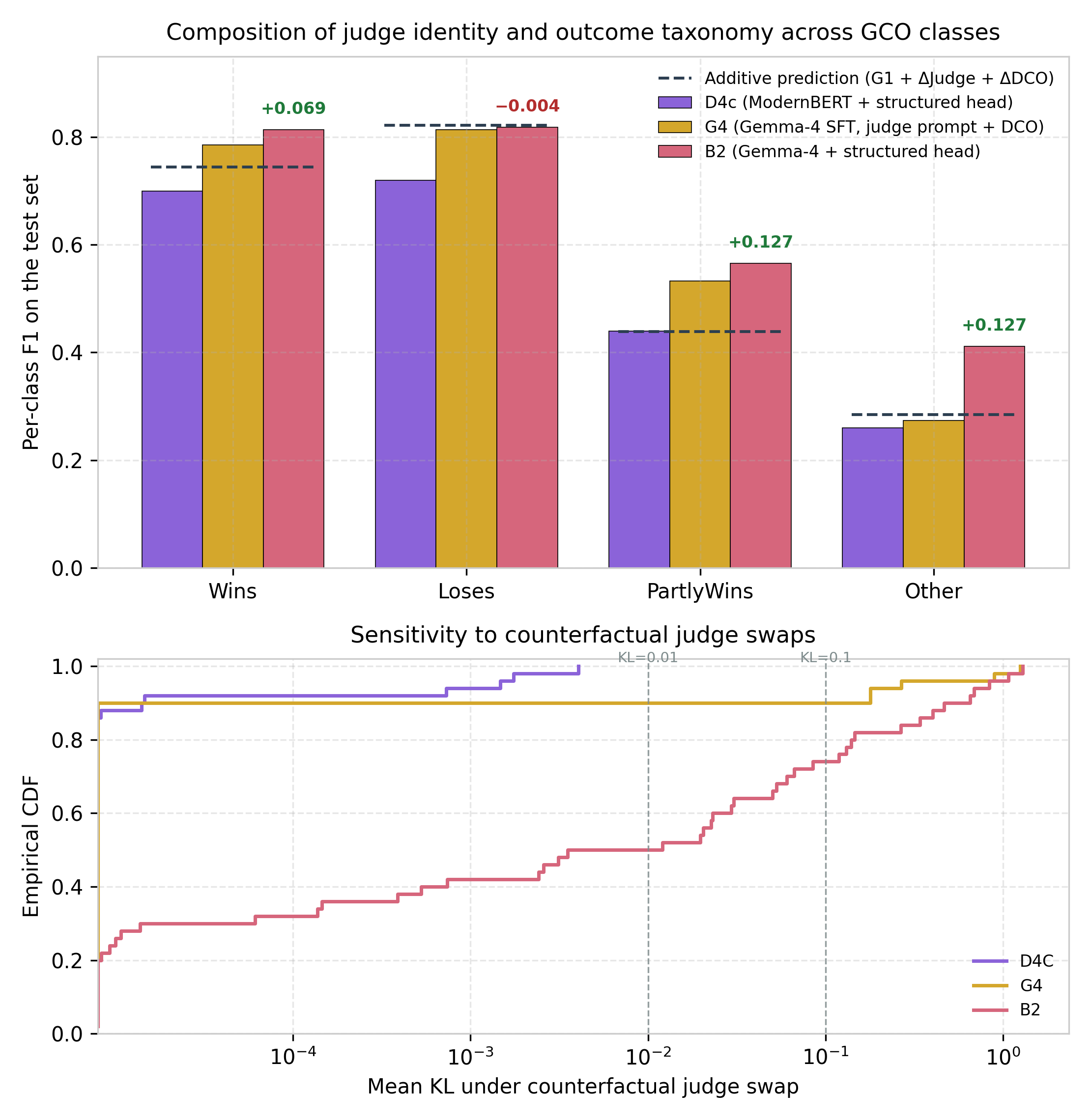}
        \caption{\textbf{B2 exceeds the additive expectation on rare/fuzzy classes and uses judge identity more actively than G4.} \textit{Top:} per-class GCO F1 with dashed markers for the additive G-track expectation; B2's excess concentrates on \textit{PartlyWins} and \textit{Other}. \textit{Bottom:} empirical CDF of mean KL divergence under counterfactual judge swaps (mean KL $0.141$ for B2 vs $0.055$ for G4).}
        \label{fig:compositional}
    \end{figure}

\paragraph{Per-class decomposition.}
Figure~\ref{fig:compositional} (top) and Table~\ref{tab:perclass} report per-class GCO F1 for the headline models. B2's largest gains over the strongest G-track baseline G3 fall on the rare and fuzzy classes (\textit{PartlyWins}, \textit{Other}); the dashed markers in Figure~\ref{fig:compositional} (top) show that B2 exceeds the additive G-track expectation (G1 plus the marginal lifts from judge prompting and DCO co-prediction) by $+0.127$ F1 on each. Appendix~\ref{sec:appendix_g4_b2_rescues} traces this aggregate pattern to individual cases where G4 predicts the wrong coarse outcome but B2 recovers a legally meaningful mixed or atypical outcome (e.g.~a security-clearance and reasonable-adjustments dispute rescued as \texttt{PARTLY\_WINS}, a protected-act dismissal dispute rescued as \texttt{OTHER}).

\paragraph{Counterfactual judge-swap intervention.}
To quantify how strongly each model conditions on judge identity at inference time, we apply a controlled intervention on the test set. For every test case $(x, j)$ we hold the case text $x$ fixed and replace the actual presiding judge $j$ with $k{=}32$ alternatives drawn uniformly from the training-judge pool ($\geq{}10$ train cases each, $j$ excluded), recompute the model's GCO distribution under each substitution, and record the Kullback-Leibler (KL) divergence, $\mathrm{KL}\!\left(p_\theta(y\mid x, j)\,\|\,p_\theta(y\mid x, j')\right)$. The per-case mean $\bar{\Delta}(x)$ over the $k$ swaps is plotted as an empirical CDF in Figure~\ref{fig:compositional} (bottom): mass near zero indicates the prediction is invariant to judge identity, mass at large $\bar{\Delta}$ indicates active judge conditioning. The intervention is identical across models, so the comparison isolates the conditioning interface.

\begin{table}[t]
    \centering
    \scalebox{0.85}{
    \begin{tabular}{lcccc}
    \hline
    \textbf{Class} & \textbf{G3} & \textbf{G4} & \textbf{B1} & \textbf{B2} \\ \hline
    Wins ($n{=}715$) & 79.40 & 78.50 & 75.60 & \textbf{81.40} \\
    Loses ($n{=}923$) & \textbf{82.00} & 81.40 & 76.80 & 81.80 \\
    PartlyWins ($n{=}370$) & 46.50 & 53.30 & 48.00 & \textbf{56.60} \\
    Other ($n{=}83$) & 32.40 & 27.30 & 31.80 & \textbf{41.10} \\ \hline
    \end{tabular}}
    \caption{Per-class GCO F1 on the UKET test set ($n{=}2091$). B2 improves all four classes relative to G3 with gains concentrated on the rare/fuzzy classes \textit{PartlyWins} and \textit{Other}.}
    \label{tab:perclass}
    \end{table}
    
\begin{figure}[t]
    \centering
    \includegraphics[width=\linewidth]{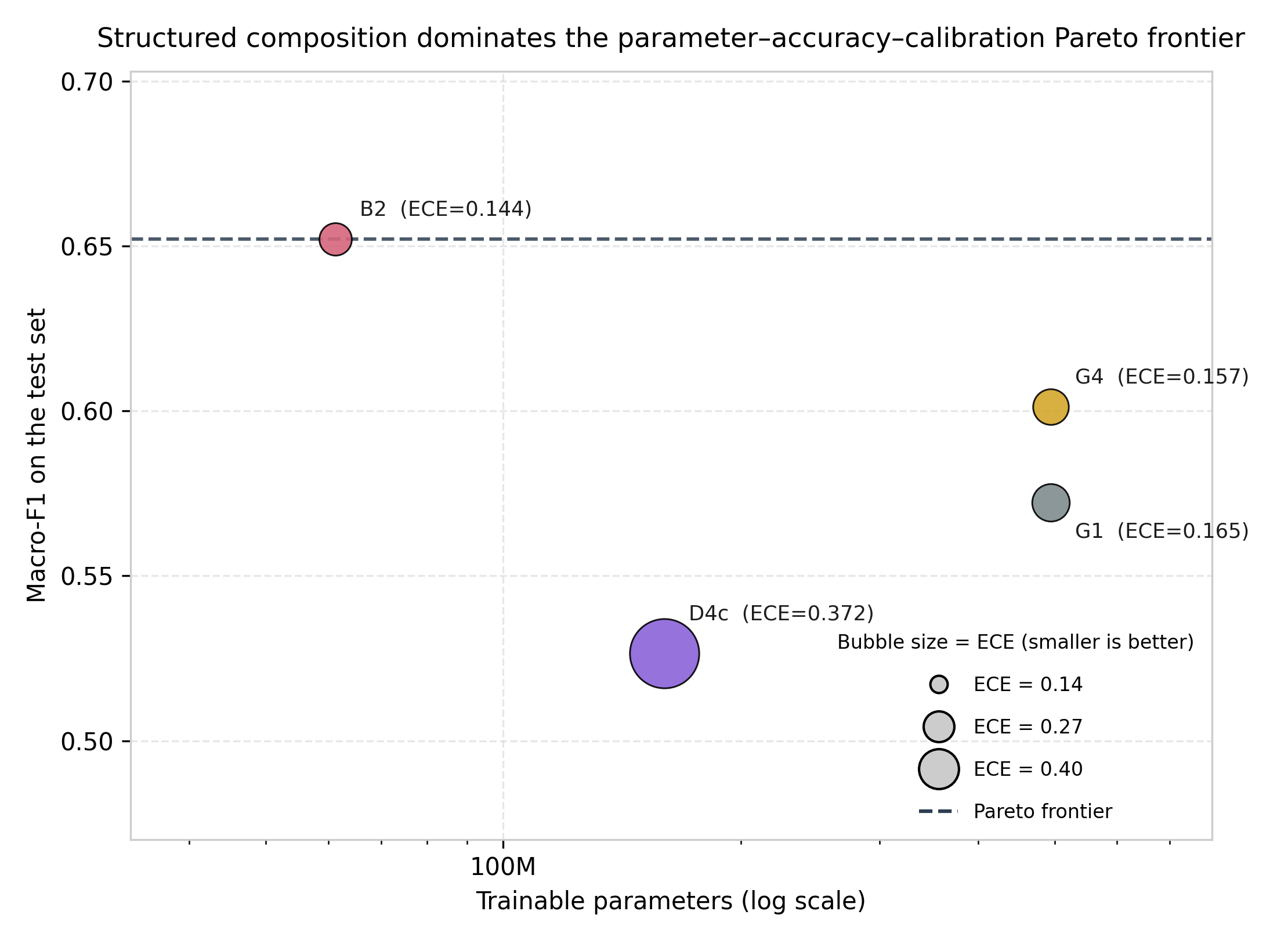}
    \caption{\textbf{Structured composition dominates the parameter--accuracy--calibration Pareto frontier.} Macro-F1 vs.\ trainable parameters (log scale); bubble size encodes ECE (larger is worse). B2 leads on all three axes.}
    \label{fig:efficiency}
\end{figure}

\paragraph{Parameter efficiency and calibration.}
Structured composition strictly dominates the parameter-accuracy-calibration Pareto frontier (Figure~\ref{fig:efficiency}). By routing signals through a compact, dedicated head, the hybrid architecture (B2) establishes a new frontier while training only a fraction of the parameters required by generative fine-tuning. Furthermore, it yields the best expected calibration error among the top-performing models. We report top-label Expected Calibration Error (ECE) with $10$ equal-width confidence bins on the maximum softmax probability, computed as the test-set weighted mean of $|\mathrm{acc}_b - \mathrm{conf}_b|$ across bins. This demonstrates that preserving explicit architectural pathways is fundamentally more efficient---and more reliable---than relying entirely on large-scale autoregressive generation.

\subsection{Latent Structure of Judicial Behavior}

To determine if the learned judge embeddings from the D4c variant capture meaningful behavior, we projected their 48 dimensions into 2D using t-SNE and applied $k$-means clustering ($k=8$). We interpreted these clusters via a ``relative specialization profile''---the difference between a cluster's average gate activation and the global average.

\begin{figure}[ht]
    \centering
    % Ensure the filename matches your saved normalized plot
    \includegraphics[width=0.50\textwidth]{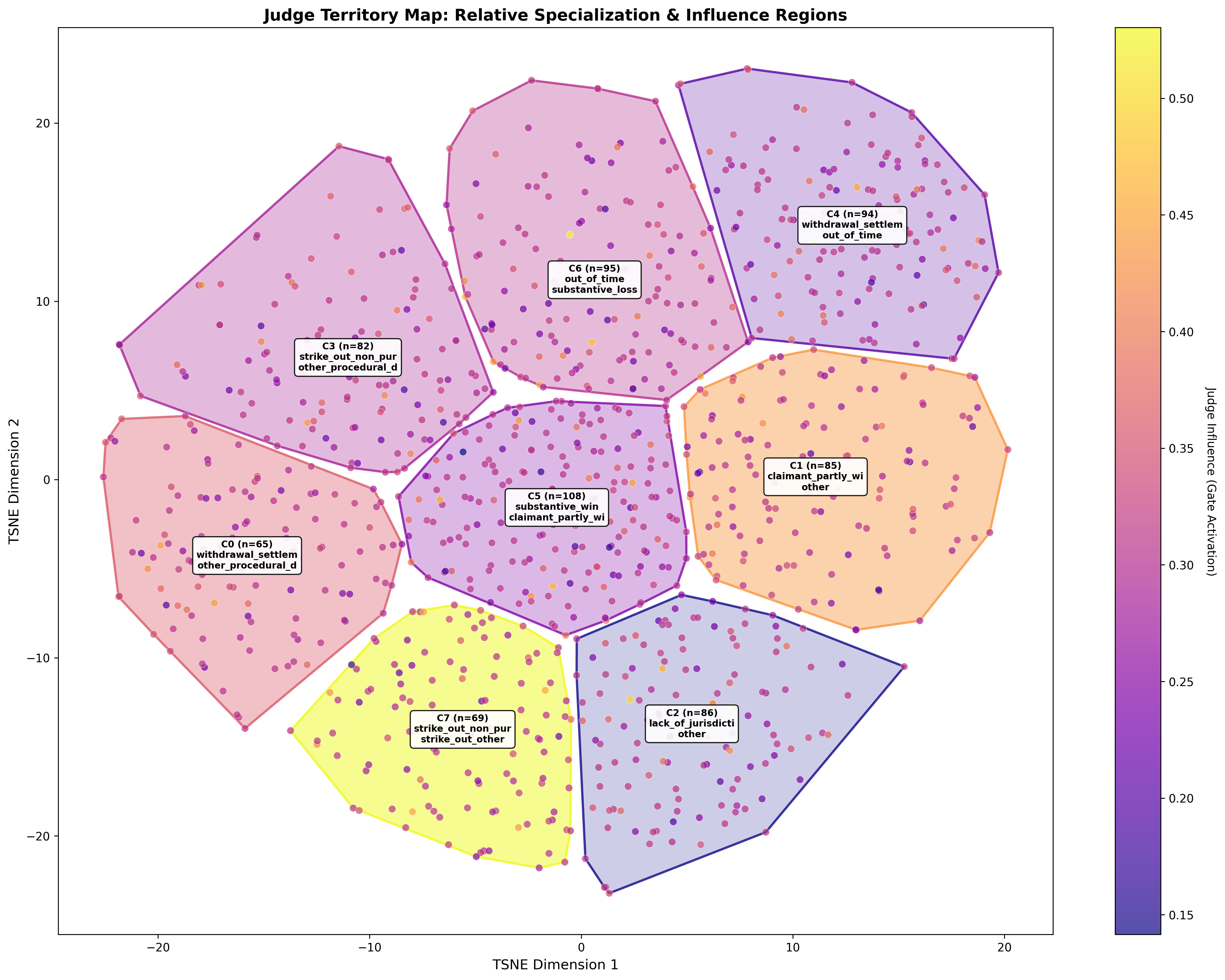}
    \caption{\textbf{Judge Territory Map: Latent Specialization and Influence.} t-SNE projection of judge embeddings with behavioral clusters ($k=8$, shaded regions). Annotations indicate the top-2 outcomes with the highest relative lift per cluster. The color gradient denotes average gate activation (Judge Influence).}
    \label{fig:judge_map}
\end{figure}

The resulting ``Judge Territory Map'' (Figure~\ref{fig:judge_map}) shows that the model learns a semantically structured representation of judicial behavior rather than treating judge identity as a random label. The embeddings form coherent regions aligned with distinct legal functions: a clear geometric separation between judges who rule on procedural grounds (e.g., Cluster 0: \textit{Default Judgment}, \textit{Lack of Jurisdiction}) and those who decide on the merits (e.g., Cluster 1: \textit{Substantive Win/Loss}); finer-grained clusters capturing specific case-management styles such as \textit{Strike Out (Non-Pursuit)} and \textit{Withdrawal}; and a gate-activation gradient under which judges in merit-based regions trigger systematically higher gate activations than those in non-merit regions, so the model learns to weigh judicial identity more heavily for complex fact-dependent outcomes than for rigid non-merit defaults.

\section{Conclusion}

In this work, we demonstrated that robust legal outcome prediction requires explicitly disentangling objective case facts from adjudicative variance. To achieve this, we introduced a Judge-Aware Gated Multi-Task architecture that dynamically modulates the influence of judicial identity based on the merit or non-merit-based nature of the dispute. By comparing a ModernBERT-based discriminative architecture, generative supervised fine-tuning of a Gemma-4 26B-A4B backbone, and a hybrid that re-uses the structured discriminative head on top of the same Gemma-4 encoder, we isolate the \emph{conditioning interface} from backbone scale. Prompt-level composition of judge identity and fine-grained outcome supervision is sub-additive, whereas routing the same signals through a learned judge embedding and a parallel multi-task head restores additive composition: the hybrid B2 reaches $65.21$ macro-F1, with gains concentrated on the rare/fuzzy classes \emph{PartlyWins} and \emph{Other} where structured composition should matter most. These findings suggest a dynamic that likely extends beyond the UKET jurisdiction: when an identity variable must shape representations through the task gradient, encoding it as a learned parameter and fusing it via structured attention can outperform encoding it as a prompt token, even at substantially larger backbone scale --- and the same gradient-accessible embeddings that drive accuracy also yield the interpretable surfaces that make the reasoning process auditable by design.
 
More broadly, this highlights a potential design heuristic for parameter-efficient fine-tuning under structural priors: variables whose effects must be rigorously controlled, audited, or shaped through supervision may be better modeled as learnable parameters exposed to the gradient, rather than being absorbed into the input distribution. Under this separation of concerns, the backbone supplies general representational capacity while a compact structured head carries the task-specific inductive bias, which explains why a single architectural choice can simultaneously improve accuracy, calibration, and trainable-parameter efficiency rather than trading among them. Identity-conditioned hierarchical classification recurs across domains such as personalization, demographic-aware modeling, and clinical decision support; we hypothesize that similar compositional bottlenecks in prompt-routed conditioning --- and similar gradient-routed remedies --- may apply in those settings. Read against the broader scale-versus-structure debate, our findings are a small but concrete instance of the argument that, in domains with strong compositional or institutional structure, choosing the right inductive bias remains complementary to gains in language-model scale.

\section*{Limitations}

While our framework achieves state-of-the-art performance and offers new interpretability into adjudicative variance, our findings should be interpreted in light of the following limitations.

A primary constraint, shared with other benchmarks derived from the Cambridge Law Corpus \cite{xie2024clcuketdatasetbenchmarkingcase}, is that the inputs are extracted from final written judgments rather than original case filings, which are not available at scale. Because these summaries are authored retrospectively, they may contain implicit "post-hoc rationalizations" where the factual narrative is unconsciously structured to support the decision. Consequently, while our model successfully distinguishes substantive from procedural signals, some predictive performance may stem from detecting these linguistic cues rather than modeling the raw legal merit. However, this limitation underscores the utility of our Gated Fusion approach: by explicitly modeling judicial identity, our architecture helps disentangle a judge's specific writing style—captured in the embedding—from the objective legal facts.

Our Judge-Aware Gated Fusion mechanism learns judicial representations directly from outcome data. While the geometric clustering of these embeddings reveals meaningful "merit-based" versus "non-merit-based" archetypes, these embeddings are data-driven proxies for behavior, not psychological profiles. They capture correlations in decision-making patterns within the available data but do not explain the causal internal reasoning of specific judges. Furthermore, our filtering strategy required judges to have a minimum of two associated cases to initialize stable embeddings. Consequently, our model’s ability to characterize adjudicators with extremely sparse histories remains limited compared to those with extensive case logs.

The DCO taxonomy, which serves as the foundation for our multi-task supervision, was generated via an LLM-assisted pipeline. While this granular schema (distinguishing, for instance, between Strike Out and Default Judgment) provides essential regularization, it acts as "silver-standard" supervision. The reliance on automated annotation introduces a potential noise floor that purely human-annotated taxonomies could avoid.

Our G-track and B-track findings are also evaluated on a single LLM family (Gemma-4 26B-A4B) and a single seed, with LoRA-only fine-tuning rather than full SFT for the G-track; while the diagnostic variants G4-H1 and G4-H2 rule out the most obvious capacity and formatting confounds, full fine-tuning of the same backbone or transfer to other LLM families (e.g.\ Llama, Qwen) may compose differently. The B1 layer choice ($L_{15}$) was selected via a small pilot sweep on Gemma-4 specifically and need not transfer.

Finally, a notable limitation of our empirical evaluation is its reliance on a single dataset. While evaluating across multiple datasets is the standard ideal in machine learning, Legal Judgment Prediction (LJP) is inherently jurisdiction-specific. The UKET dataset is uniquely suited for this study on judicial discretion because it preserves judge identities. Many other jurisdictions systematically redact judge identities in published rulings, which prevents the scalable analysis required to train and evaluate our Judge-Aware architecture. Consequently, while our study incorporates conceptually and technically relevant advancements from the broader LJP literature, our architectural innovations are necessarily tailored to the specific nature of the UK Employment Tribunal, where procedural disposals are frequent and decisive. We emphasize that our classification is functional and approximate, intended to capture whether the judge engages with case facts, rather than serving as a rigorous universal doctrinal taxonomy.

Future work is required to validate whether this framework and its gating mechanism can be adapted to other courts where judicial discretion manifests differently—such as through sentencing severity or damages awarded—provided that the requisite judge identities are accessible.

\section*{Impact Statement}

The curated dataset is developed on the basis of the CLC-UKET corpus \cite{ostling2024cambridgelawcorpusdataset}, which aggregates publicly available UK legal judgments derived from the Cambridge Law Corpus (CLC). Both the decisions in the CLC and the jurisdiction codes of UKET are licensed for use under the Open Government Licence. This licence grants a worldwide, royalty-free, perpetual, and non-exclusive licence. Access to the CLC is restricted to researchers with confirmed ethical clearance and requires compliance with the DPA and UK GDPR. Whilst UK legal judgments are not anonymised, Rule 49 of the Employment Tribunal Rules 2024 ensures that sensitive personal information is anonymised when necessary. Additionally, Schedule 2, Part 5 of the DPA provides derogations for academic research, alleviating the burden of notifying all individuals involved in judgments.

A distinct ethical consideration in this work concerns the explicit modeling of judicial identity to quantify discretionary variance. We are acutely aware of the legal and ethical sensitivities surrounding judicial profiling, particularly given that some jurisdictions strictly prohibit the statistical analysis of judge identity for predictive purposes (e.g., France’s Art. 33, Law No. 2019-222). However, our work is situated within the English legal order, which operates fundamentally on the principle of "open justice," where judicial names and decisions are matters of public record. 

Crucially, we draw a strict ethical distinction between profiling individuals for applied litigation strategy—which raises significant ethical concerns—and the empirical analysis of decision patterns to audit systemic fairness. Our objective aligns with established empirical legal studies \citep{engel2020manna} by treating predictive systems as diagnostic instruments rather than decision-making tools. The latent "judge embeddings" and the resulting "Judge Territory Maps" are strictly statistical proxies designed to evaluate whether similar cases are treated differently across adjudicators. They are not intended, nor should they be used, as psychological profiles, performance metrics, or commercial tools for litigation support. 

Furthermore, while judge names are public, we go beyond baseline requirements by ensuring that specific judge identities are not revealed in our qualitative outputs or visualisations. This ensures our research remains firmly within GDPR privileges for scientific inquiry, auditing the structural variance in the legal system without targeting individual adjudicators.

Our dataset does not go beyond publicly available information and includes established procedures for data removal if requested. Like the original CLC, access to the dataset created for this paper is limited to qualified researchers who adhere to the relevant ethical and legal standards. For more details on the legal and ethical considerations concerning the underlying CLC dataset, see \cite{ostling2024cambridgelawcorpusdataset}.

\nocite{langley00}

\bibliography{example_paper}
\bibliographystyle{icml2026}

%%%%%%%%%%%%%%%%%%%%%%%%%%%%%%%%%%%%%%%%%%%%%%%%%%%%%%%%%%%%%%%%%%%%%%%%%%%%%%%
%%%%%%%%%%%%%%%%%%%%%%%%%%%%%%%%%%%%%%%%%%%%%%%%%%%%%%%%%%%%%%%%%%%%%%%%%%%%%%%
% APPENDIX
%%%%%%%%%%%%%%%%%%%%%%%%%%%%%%%%%%%%%%%%%%%%%%%%%%%%%%%%%%%%%%%%%%%%%%%%%%%%%%%
%%%%%%%%%%%%%%%%%%%%%%%%%%%%%%%%%%%%%%%%%%%%%%%%%%%%%%%%%%%%%%%%%%%%%%%%%%%%%%%
\newpage
\appendix
% \onecolumn
\section{DCO Taxonomy and Extraction Logic}
\label{app:dco_taxonomy}

The resulting DCO schema comprises 11 outcome categories derived from a multi-stage extraction pipeline. Raw outcomes were first classified with prompts designed to distinguish non-merit disposals from merit-based determinations.

To map these raw extraction outputs to the final labels used in our experiments, we employed a logic-based aggregation strategy. First, granular source logic labels were extracted from the case text. These were then assigned to the specific DCO labels as detailed in Table~\ref{tab:dco-aggregation}.

\begin{table}[h]
\centering
\small
\setlength{\tabcolsep}{3pt} % Tighten padding
\begin{tabular}{|p{0.35\linewidth}|p{0.6\linewidth}|}
\hline
\textbf{DCO Label} & \textbf{Constituent Source Labels / Logic} \\ \hline
substantive\_win & \textit{merit\_based} (from Wins) \\ \hline
substantive\_loss & \textit{merit\_based} (from Losses) \\ \hline
default\_judgment & \textit{default\_judgment\_no\_response}, \textit{default\_judgment\_invalid\_response} \\ \hline
claimant\_partly\_wins & Identity map on GCO \\ \hline
out\_of\_time & \textit{out\_of\_time} \\ \hline
strike\_out\_non\_pursuit & \textit{strike\_out\_non\_pursuit} \\ \hline
strike\_out\_other & \textit{strike\_out\_vexatious}, \textit{strike\_out\_non\_compliance}, \textit{strike\_out\_no\_prospect}, \textit{strike\_out\_fair\_hearing} \\ \hline
other\_procedural dismissal & \textit{non\_attendance\_claimant}, \textit{invalid\_acas\_cert}, \textit{defective\_pleadings}, \textit{other\_procedural} \\ \hline
lack\_of\_jurisdiction & \textit{lack\_of\_jurisdiction} \\ \hline
withdrawal\_settlement & \textit{withdrawal}, \textit{acas\_settlement}, \textit{private\_settlement} \\ \hline
other & Identity map on GCO \\ \hline
\end{tabular}
\caption{Aggregation logic mapping constituent source labels to final DCO classes.}
\label{tab:dco-aggregation}
\end{table}

Following this aggregation, we adopt a functional taxonomy. We distinguish between \textbf{Merit-Based} outcomes (where factual evidence or sufficiency is assessed) and \textbf{Non-Merit-Based} disposals (technical or case management terminations).

Table~\ref{tab:dco-functional} details the functional decomposition, descriptions, and class distribution for the final schema.

\begin{table}[ht]
\centering
\small
\setlength{\tabcolsep}{3pt}
% Preserving your exact column widths and vertical lines
\begin{tabular}{|p{0.15\linewidth}|p{0.25\linewidth}|r|p{0.4\linewidth}|}
\hline
\textbf{Func. Cat.} & \textbf{DCO Class} & \textbf{Count} & \textbf{Description} \\ \hline
\multirow{3}{=}{Merit-Based} & substantive win & 3,463 & Won on merits after evidence evaluation \\ \cline{2-4}
 & substantive loss & 3,341 & Lost on merits after evidence evaluation \\ \cline{2-4}
 & default judgment & 1,307 & (Sufficiency Check) \\ \hline
Mixed & claimant partly wins & 2,991 & Mixed outcome (Partial success) \\ \hline
\multirow{7}{=}{Non-Merit} & out of time & 731 & Dismissed as time-barred (Limitation) \\ \cline{2-4}
 & strike out non pursuit & 579 & Strike-out due to failure to pursue \\ \cline{2-4}
 & strike out other & 532 & Merged Strike-outs (Abuse/Compliance) \\ \cline{2-4}
 & other procedural dismissal & 398 & Other technical procedural defects \\ \cline{2-4}
 & lack of jurisdiction & 377 & Tribunal lacks jurisdiction \\ \cline{2-4}
 & withdrawal settlement & 193 & Withdrawal or explicit settlement \\ \cline{2-4}
 & other & 698 & Other miscellaneous outcomes \\ \hline
\end{tabular}
\caption{Functional classification and distribution of DCO labels.}
\label{tab:dco-functional}
\end{table}

These fine-grained DCO labels serve two purposes in our model:
(i) they define the outcome prototypes used by the label-wise attention mechanism, and
(ii) they act as auxiliary supervision targets in the multi-task learning objective.

\section{Training and Inference Details}
\label{app:training}

The introduction of label-wise attention, judge-aware fusion, and auxiliary supervision substantially increases model expressivity and, if left unchecked, encourages memorization. We therefore train MTL variants under a deliberately higher-regularization regime.

Our base MTL configuration employs LWAN with judge-aware gated fusion, an 11-class DCO auxiliary task, cosine learning-rate scheduling with warmup under end-to-end training. Concretely, the optimal configuration reduces the attention dimension from 256 to 192 while increasing friction via stronger weight decay (0.05), moderate label smoothing (0.09), and increased judge dropout (0.27). The auxiliary curriculum is applied with a relatively high weight ($\alpha=0.4$), while only a mild class-weighting signal is retained (power 0.17), avoiding the loss instability observed with aggressive reweighting. We find that stability and generalization are achieved primarily through friction rather than architectural bottlenecks.

Across both broad sweeps and targeted follow-up tuning, we observe a consistent stability pattern: the best trade-off occurs in a narrow mid-range around $\alpha\approx 0.4$ (empirically, $\alpha\in[0.38,0.40]$). This range repeatedly produced our strongest Pareto configurations (high Macro-F1 while maintaining strong accuracy), and therefore became the anchor region for later local sweeps.

In contrast, larger settings (notably $\alpha>0.45$) were consistently less reliable in our runs, with weaker generalization and less favorable main-task trade-offs. Intuitively, this matches expected multi-task behavior: if $\alpha$ is too small, auxiliary supervision is underused; if too large, the auxiliary signal dominates and can induce negative transfer to the primary task.

\paragraph{G-track (Gemma-4 generative baselines).}
All G-track variants apply LoRA at rank $r{=}16$, $\alpha{=}32$, dropout $0.05$, to all attention and MLP projections ($q$, $k$, $v$, $o$, $\mathrm{gate}$, $\mathrm{up}$, $\mathrm{down}$). The diagnostic variant G4-H1 uses $r{=}64$. All G-track models are trained for 3 epochs with a cosine schedule, batch size 2, and gradient accumulation over 8 steps.

\paragraph{B1 encoder details.}
\label{app:b1_encoder}
A pilot sweep over layers $\{L_{7}, L_{15}, L_{21}, L_{27}, L_{30}\}$ identifies $L_{15}$ as the strongest probe, consistent with the observation that MoE routing reorganises features earlier than in dense decoders. The \emph{echo embedding} readout \citep{springer2024echoembeddings} duplicates the input sequence and reads hidden states from the second copy, recovering bidirectional context without retraining the backbone. We additionally concatenate per-expert router activations (softmaxed and averaged across MoE layers) to the hidden states \citep{li2025moeembedding}.

\paragraph{B2 training.}
B2 applies LoRA ($r{=}32$, $\alpha{=}64$, dropout $0.05$) to the Gemma-4 encoder and trains it jointly with the structured head for 5 epochs, using a cosine schedule with learning rate $1\times10^{-4}$ for the encoder and $5\times10^{-5}$ for the head.

\section{Dataset Statistics and Preprocessing}
\label{app:dataset_statistics}

This section expands on the dataset filtering pipeline and judge stratification strategy introduced in Section 4.1.

Table~\ref{tab:dataset_stats} summarizes the data exclusion stages applied to the raw corpus to establish our final working dataset, along with the resulting partition sizes.

\begin{table}[ht]
\centering
\scalebox{0.9}{
\begin{tabular}{lr}
\hline
\textbf{Stage} & \textbf{Count} \\ \hline
Raw Cases Loaded & 14,610 \\
Excluded (Missing/Ambiguous Judge ID) & 673 \\
\textbf{Final Experimental Dataset} & \textbf{13,937} \\ \hline
\textit{Split Distribution} & \\
Training Set (70\%) & $\sim$9,755 \\
Validation Set (15\%) & $\sim$2,091 \\
Test Set (15\%) & $\sim$2,091 \\ \hline
\end{tabular}}
\caption{Dataset statistics and filtering pipeline.}
\label{tab:dataset_stats}
\end{table}

\textbf{Data Splitting and Judge Stratification.} The data was partitioned into training (70\%), validation (15\%), and testing (15\%) sets. A rigorous ``Judge-Aware'' splitting strategy was employed to prevent data leakage and test true generalization. The training set was filtered to include only judges with a minimum of 2 associated cases to ensure stable embedding initialization. The validation and test sets retained all judges, including ``unseen'' or rare judges who did not appear in the training set to accurately reflect real-world inference conditions.

Table~\ref{tab:dataset_statistics} reports the detailed token-level statistics for the \texttt{facts} and \texttt{claims} fields across these exact experimental splits.

\begin{table}[ht]
\centering
\begin{tabular}{lccc}
\hline
 & \textbf{train} & \textbf{val} & \textbf{test} \\
\hline
\#Cases & 9,755 & 2,091 & 2,091 \\
\#AvgFactLen & 79 & 79 & 80 \\
\#MaxFactLen & 408 & 273 & 426 \\
\#AvgClaimLen & 32 & 33 & 32 \\
\#MaxClaimLen & 186 & 142 & 163 \\
\hline
\end{tabular}
\caption{Statistics of the experimental splits. \#Cases denotes the number of cases. \#AvgFactLen and \#AvgClaimLen denote the average number of words in the \texttt{facts} and \texttt{claims} fields, respectively. \#MaxFactLen and \#MaxClaimLen denote the corresponding maximum lengths.}
\label{tab:dataset_statistics}
\end{table}

\section{Gate Influence Analysis}
\label{app:gate_influence}

\begin{figure}[ht]
    \centering
    \includegraphics[width=0.8\linewidth]{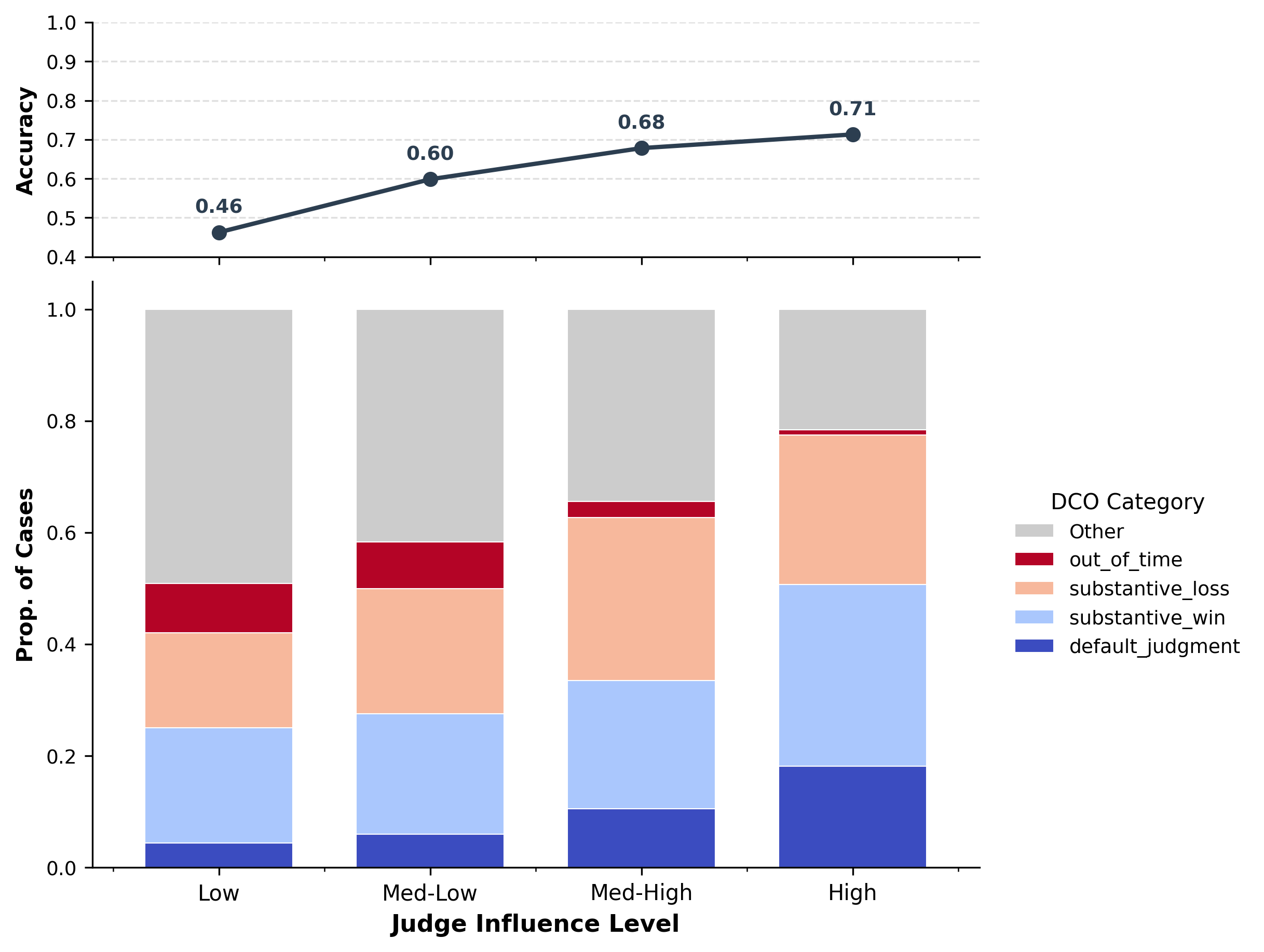}
    \caption{Impact of judge-gate influence on model accuracy and DCO composition in D4c. \textit{Top:} overall accuracy increases monotonically with gate activation quartile. \textit{Bottom:} DCO distribution shifts across quartiles; categories with high variance are shown individually, stable or rare categories are aggregated as ``Other''.}
    \label{fig:influence_analysis}
\end{figure}

To quantify the semantic role of the judge-aware gate in D4c variant, we stratify the test set into four quartiles by average gate activation magnitude (Low, Med-Low, Med-High, High). Figure~\ref{fig:influence_analysis} shows the combined analysis. The top panel illustrates a positive correlation between gate activation and main-task accuracy: the model relies more heavily on the judge embedding precisely on those cases where it achieves higher performance. The bottom panel shows the changing DCO composition across quartiles: \textit{strike\_out\_non\_pursuit} is concentrated in the low-influence region and vanishes in the highest quartile, indicating that straightforward technical terminations are resolved from text patterns alone. Conversely, \textit{default\_judgment} expands dramatically in the highest quartile, consistent with the legal intuition that default-judgment rulings require the model to identify a specific cluster of strict proceduralist judges rather than relying on case-text content.

\section{Qualitative case studies: G4--B2 rescue cases}
\label{sec:appendix_g4_b2_rescues}

To complement the aggregate results, we examine individual cases where G4 is
incorrect but B2 predicts the coarse outcome correctly. These are representative
rescue cases drawn from the 199-case set satisfying
\texttt{G4.pred\_gco} $\neq$ gold and \texttt{B2.pred\_gco} = gold.

Both G4 and B2 use the Gemma backbone, but they expose the legal signals
differently. G4 must express judge context and fine-grained outcome information
through a prompt and a single autoregressive output channel. B2 routes the same
backbone representation through a composite head consisting of LWAN, judge-gated fusion, and a multi-task classifier. The cases
below illustrate where that composite head changes the prediction.

% -----------------------------------------------------------------------
%  CASE 1 -- security clearance, reasonable adjustments, partial success
% -----------------------------------------------------------------------
\subsection{Rescuing a partial-win outcome in a security-clearance and reasonable-adjustments dispute}
\label{subsec:g4_b2_case_security_clearance}

Table~\ref{tab:g4_b2_case_security_clearance} shows a public-sector recruitment
case involving security-clearance requirements, conflicting administrative
communications, and disability-related reasonable-adjustment claims. G4 predicts
a complete claimant loss and maps the detailed outcome to
\texttt{substantive\_loss}. B2 instead predicts the correct coarse label,
\texttt{PARTLY\_WINS}.

\begin{table}[ht]
\centering
\footnotesize
\renewcommand{\arraystretch}{1.15}
\setlength{\tabcolsep}{6pt}
\begin{tabular}{@{}l l@{}}
\hline
\textbf{Source} & \textbf{GCO} \\
\hline
Gold & \texttt{PARTLY\_WINS} \\
G4   & \texttt{LOSES} $\times$ \\
B2   & \texttt{PARTLY\_WINS} \checkmark \\
\hline
\end{tabular}
\caption{Security-clearance rescue case; learned judge cluster C6). G4
collapses a mixed recruitment and reasonable-adjustments dispute to claimant
loss; B2 recovers the partial-win coarse outcome.}
\label{tab:g4_b2_case_security_clearance}
\end{table}

\paragraph{Case text (anonymized).}

\begin{quote}
\small
\textit{``[Claimant] applied for two public-sector roles and was successful in
obtaining both. The roles required security clearance. [Claimant] was initially
asked to complete one level of security questionnaire, but believed that a
different counterterrorist-level check was required. [Claimant] repeatedly raised the
issue with the respondent and received conflicting information before being told
that an error had been made. [Claimant] also identified anxiety and depression
as disabilities and requested reasonable adjustments in the recruitment process,
arguing that delays and unclear email communications exacerbated those
conditions. The claims included race discrimination and disability
discrimination, including alleged failures to make reasonable adjustments.''}
\end{quote}

Key observations:

\begin{enumerate}
    \item \textbf{Legal evaluation.}
    The dispute combines administrative process, security vetting, race
    discrimination, disability discrimination, and reasonable adjustments. The
    facts do not point cleanly to a complete win or complete loss. This is the
    type of mixed legal evaluation that is naturally represented by
    \texttt{claimant\_partly\_wins}.

    \item \textbf{G4 collapses the mixed evaluation to a full loss.}
    The generative model predicts both \texttt{LOSES} and
    \texttt{substantive\_loss}. That error is plausible if the model focuses on
    the contested security-clearance dispute as an unsuccessful discrimination
    claim, but it misses the multiple remedy structure of the case.

    \item \textbf{B2 recovers the partial-outcome structure.}
    B2 predicts \texttt{PARTLY\_WINS}, aligning with the gold GCO. This is
    consistent with the aggregate result that B2's largest gains are on
    \texttt{PARTLY\_WINS}. The composite head can preserve a fine-grained
    partial-success pathway that is easy to lose in a single autoregressive
    label channel.

    \item \textbf{Implication for the conditioning interface.}
    Since both models use the Gemma backbone, the relevant difference is the
    conditioning interface. The rescue suggests that LWAN plus multi-task
    fine-grained supervision helps the model retain mixed remedial structure
    even when the surface narrative contains strong loss-like elements.
\end{enumerate}

% -----------------------------------------------------------------------
%  CASE 2 -- protected act, dismissal, atypical OTHER outcome
% -----------------------------------------------------------------------
\subsection{Rescuing an atypical outcome in a protected-act dismissal dispute}
\label{subsec:g4_b2_case_protected_act}

Table~\ref{tab:g4_b2_case_protected_act} shows a dismissal dispute involving an
alleged protected act, pupil safeguarding concerns, possible data-breach issues,
and loss of trust and confidence. G4 predicts the majority-style
\texttt{LOSES}/\texttt{substantive\_loss} pathway, while B2 predicts the correct
\texttt{OTHER} outcome.

\begin{table}[ht]
\centering
\footnotesize
\renewcommand{\arraystretch}{1.15}
\setlength{\tabcolsep}{6pt}
\begin{tabular}{@{}l l@{}}
\hline
\textbf{Source} & \textbf{GCO} \\
\hline
Gold & \texttt{OTHER} \\
G4   & \texttt{LOSES} $\times$ \\
B2   & \texttt{OTHER} \checkmark \\
\hline
\end{tabular}
\caption{Protected-act rescue case; learned judge cluster C3). G4 maps a
procedurally atypical dismissal dispute to the ordinary substantive-loss
pathway; B2 recovers the minority \texttt{OTHER} outcome.}
\label{tab:g4_b2_case_protected_act}
\end{table}

\paragraph{Case text (anonymized).}

\begin{quote}
\small
\textit{``[Claimant] was employed as a sixth-form pastoral leader and was
dismissed for alleged gross misconduct. He alleged that the real reason for
dismissal was that he had undertaken a protected act by raising concerns that the
respondent was not observing its duty of care towards a pupil with a mental
impairment. The respondent argued that [Claimant] had engaged in inappropriate
communications with the pupil and the pupil's parent, failed to disclose details
of a potential data breach, failed to cooperate with the investigation, and
covertly recorded an investigation meeting. The claims included victimisation
under the Equality Act and sex discrimination.''}
\end{quote}

Key observations:

\begin{enumerate}
    \item \textbf{Legal evaluation.}
    The case lies at the boundary between ordinary dismissal litigation,
    victimisation for a protected act, safeguarding concerns, data governance,
    and investigation conduct. This combination is not well captured by a simple
    merits-based win/loss framing.

    \item \textbf{G4 follows the dominant substantive-loss pathway.}
    Given the misconduct narrative and the employer's loss-of-trust account, a
    generative model can plausibly map the case to \texttt{LOSES}. But the gold
    label is \texttt{OTHER}, reflecting the atypical procedural and legal
    evaluation. In particular, further evidence might be necessary in this case.

    \item \textbf{B2 preserves the minority-class boundary.}
    B2 predicts \texttt{OTHER}, the rarest and most difficult GCO class. This
    matches the aggregate pattern: B2 improves Other F1 to 0.411, compared with
    G4's 0.273.

    \item \textbf{Implication for the composite head.}
    The case illustrates a central advantage of the composite head:
    label-wise attention and fine-grained supervision can maintain an
    alternative procedural/outcome pathway even when the surface text contains
    strong majority-class cues.
\end{enumerate}

%%%%%%%%%%%%%%%%%%%%%%%%%%%%%%%%%%%%%%%%%%%%%%%%%%%%%%%%%%%%%%%%%%%%%%%%%%%%%%%
%%%%%%%%%%%%%%%%%%%%%%%%%%%%%%%%%%%%%%%%%%%%%%%%%%%%%%%%%%%%%%%%%%%%%%%%%%%%%%%

\end{document}